\documentclass[conference]{IEEEtran}
\IEEEoverridecommandlockouts
\usepackage{cite}
\usepackage{amsmath,amssymb,amsfonts}
\usepackage{algorithmic}
\usepackage{graphicx}
\usepackage{textcomp}
\usepackage{xcolor}
\usepackage{booktabs}
\usepackage{multirow}
\usepackage{arydshln}

\newcommand\fig[1]{\textbf{Fig.}~\ref{#1}}

\newcommand\tab[1]{\textbf{Tab.}~\ref{#1}}

\def\BibTeX{{\rm B\kern-.05em{\sc i\kern-.025em b}\kern-.08em
    T\kern-.1667em\lower.7ex\hbox{E}\kern-.125emX}}

\usepackage{etoolbox}
\patchcmd{\abstract}{\vspace*{12pt}}{\vspace*{0pt}}{}{}
\begin{document}

\title{Late Breaking Results: Quamba-SE:
 Soft-edge  Quantizer for Activations  in State Space Models}

\author{\IEEEauthorblockN{Yizhi Chen \{yizhic@kth.se\} and Ahmed Hemani \{hemani@kth.se\} }
\IEEEauthorblockA{Department of Electronics and Embedded Systems,
KTH Royal Institute of Technology, Stockholm, Sweden }}

\maketitle
\begin{abstract}
We propose Quamba-SE, a soft-edge quantizer for State Space Model (SSM) activation quantization. Unlike existing methods, using standard INT8 operation, Quamba-SE employs three adaptive scales: high-precision for small values, standard scale for normal values, and low-precision for outliers. This preserves outlier information instead of hard clipping, while maintaining precision for other values. We evaluate on Mamba-130M across 6 zero-shot benchmarks. Results show that Quamba-SE consistently outperforms Quamba, achieving up to +2.68\% on individual benchmarks and up to +0.83\% improvement in the average accuracy of 6 datasets.
\end{abstract}


\section{Introduction}
Large Language Models (LLMs) are rapidly evolving, dominated by Transformer architectures like GPT. Recently, State Space Models (SSMs), such as Mamba \cite{gu2024mamba}, have emerged as significant alternatives. Chiang \textit{et al.} \cite{chiangquamba} show that Mamba series outperform Pythia \cite{biderman2023pythia}, the same-sized Transformers.

Quantization is crucial for enhancing speed and reducing storage. For LLMs, Post-Training Quantization (PTQ) \cite{pierro2024mamba} is widely used as QAT (Quantization-Aware-Training) \cite{liu2024llm} requires significant computational training costs.  However, related works  \cite{pierro2024mamba,chiangquamba,chiangquamba2} reveal unique challenges in post-training quantization: SSM activations contain significant outliers. Although outliers constitute a small percentage, including them damages quantization precision for normal values.

Existing methods include Hadamard Transform \cite{chiangquamba,chiangquamba2,xumambaquant}, percentile clipping \cite{chiangquamba}, and group-wise scaling \cite{chiangquamba2}. However, these are CUDA-dependent and lack hardware-level optimization.  These methods utilize standard INT8 operations, which have only one scale for all data and hard clip outliers.

\begin{figure}[htb]
    \centering
    \includegraphics[width=0.4\textwidth,  trim={0cm 3mm 3mm 3mm}, clip]{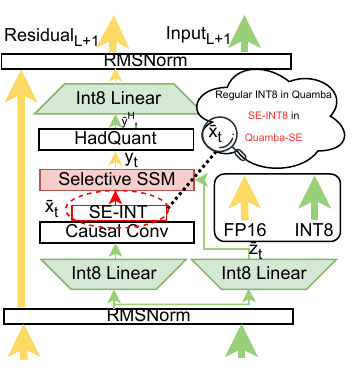}
    \vspace{-3mm}  
    \caption{Quamba-SE: dataflow and data precision}
    \label{fig:QuambaSELocation.pdf}
    \vspace{-5mm}
\end{figure}

\begin{table*}[htb]

\centering
\caption{Zero-shot accuracy (\%) comparison on Mamba-130M. Best in \textbf{bold}, second best \underline{underlined} among quantized models. $^{\dagger}$Avg. over selected benchmarks: LAMBADA, ARC-C, WinoGrande.}
\label{table:datasetresults}
\resizebox{0.95\textwidth}{!}{%
\begin{tabular}{ll|cccccc|cc|cc}
\toprule
\textbf{Setting} & \textbf{Method} & LAMBADA & HellaSwag & PIQA & ARC-E & ARC-C & WinoGrande & \textbf{Avg.} & $\Delta$ & \textbf{Avg.$^{\dagger}$} & $\Delta^{\dagger}$ \\
\midrule
\multirow{2}{*}{Reported \cite{chiangquamba}}
 & FP16 & 44.20 & 35.30 & 64.50 & 48.00 & 24.30 & 51.90 & 44.70 & -- & 40.13 & -- \\
 & Quamba & 40.60 & 35.00 & 63.00 & 46.50 & 23.00 & 53.10 & 43.50 & -- & 38.90 & -- \\
\midrule
\midrule
FP16 & FP16 & 44.24 & 35.22 & 64.57 & 48.05 & 24.35 & 52.00 & 44.74 & -- & 40.20 & -- \\
\hdashline
\multirow{2}{*}{\shortstack[l]{Calib.\\(99.99\%)}}
 & Quamba & 42.44 & \underline{35.30} & \underline{63.04} & \underline{45.92} & 23.98 & \underline{52.47} & \underline{43.86} & -- & 39.63 & -- \\
 & Quamba-SE & \underline{43.42} & 34.90 & 62.84 & \textbf{46.80} & \underline{25.05} & 51.51 & \textbf{44.08} & +0.22 & \underline{39.99} & +0.36 \\
\hdashline
\multirow{2}{*}{\shortstack[l]{Calib.\\(99.999\%)}}
 & Quamba & 42.23 & 34.69 & 62.16 & 42.23 & 24.18 & 52.45 & 42.99 & -- & 39.62 & -- \\
 & Quamba-SE & \textbf{44.56} & 35.03 & \textbf{63.06} & 42.44 & 24.37 & 52.03 & 43.58 & +0.59 & \textbf{40.32} & +0.70 \\
\hdashline
\multirow{2}{*}{\shortstack[l]{Official\\Weights \cite{quamba2025hf}}}
 & Quamba & 39.06 & 35.17 & 61.94 & 45.13 & 24.90 & 50.73 & 42.82 & -- & 38.23 & -- \\
 & Quamba-SE & 40.40 & \textbf{35.54} & 62.08 & 45.13 & \textbf{25.32} & \textbf{53.41} & 43.65 & +0.83 & 39.71 & +1.48 \\
\bottomrule
\end{tabular}%
}
 \vspace{-3mm}
\end{table*}

This is also a very new question, Quamba \cite{chiangquamba}  and Quamba2 \cite{chiangquamba2} are published in conferences in April 2025 and July 2025, respectively.  We propose Quamba-SE, a soft-edge quantizer that operates at the hardware level, providing soft edges for values instead of hard clipping. While evaluated on Quamba, the quantizer design applies to any State Space Models.
\vspace{-3mm}
\section{Background and related work}
\subsection{Background}
We illustrate Quamba and Quamba-SE's dataflow in \fig{fig:QuambaSELocation.pdf}.  Some components can be easily quantized, such as the INT8 linear projection. The SSM input and output are key challenges for quantization. For the output Y, Hadamard transforms are employed to smooth the outlier distribution.

\subsection{Related work}

We focus on the SSM input $X_t$ (highlighted in \fig{fig:QuambaSELocation.pdf}). Mamba-PTQ \cite{pierro2024mamba} reports that the challenge of quantizing SSMs stems from activation outliers. Quamba \cite{chiangquamba} uses calibration, including percentile clipping, to exclude outliers. The stored scales are employed for regular INT8 quantizations during inference, achieving higher performance than Mamba-PTQ. Quamba2 \cite{chiangquamba2} extends Quamba with group-wise scales, improving accuracy in Mamba2 \cite{dao2024Mamba2} models but showing minor improvement in Mamba1 models. Furthermore, Quamba and Quamba2, constrained by CUDA and PyTorch frameworks, focus on finding better scales for standard INT8 operations.

\section{Soft-edge Quantization Methods}

Same as Quamba and Quamba2, we calibrate and save the scale offline. The difference is during inference as highlighted in red in \fig{fig:QuambaSELocation.pdf}: Quamba uses a single scale and clips outliers, while we apply three scales adaptively during inference.
\begin{figure}[htb]
    \centering
    \includegraphics[width=0.5\textwidth, trim={0 0mm 0 0mm}, clip]{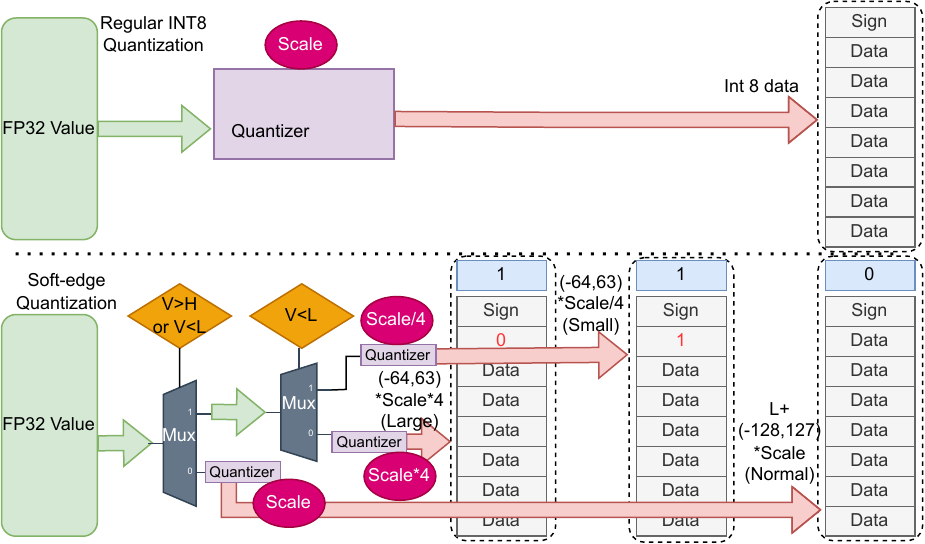}
    \caption{Hardware architecture of Quamba-SE}
    \label{fig:quambaSEHardwareArchitecture}
    \vspace{-5mm}
\end{figure}

We introduce two extra scales in \fig{fig:quambaSEHardwareArchitecture}, in addition to using the same stored scale for normal values: a high-precision scale (scale/4) for small values, a standard scale for medium values, and a low-precision scale (scale×4) for outliers.  During inference, values are first classified: normal values use standard INT8 with the soft-edge (SE) identifier disabled.  For values smaller than threshold $L$ or larger than threshold $H$, the SE identifier is enabled. To avoid extra storage, we use the second bit of INT8 data to distinguish small from large values, leaving  6 bits for special-range data. Results show that 6-bit precision for special regions is sufficient. Our solution provides superior dynamic range: high precision for small values, same precision as Quamba without any loss for medium values, and low precision (instead of clipping) for outliers.

In conclusion, while Quamba provides an effective scale, it employs standard INT8 quantization. We adopt Quamba's effective scale and design a specialized hardware quantizer to provide soft edges for different value ranges.

\section{Experiments}
We conduct our experiments on the open-source Quamba codebase \cite{quambaCodebase}  using an RTX 5090 with CUDA 12.8.  As the framework does not support customized INT8, we limit FP32 values to be the data Quamba-SE can represent, to simulate our customized Soft-edge quantizer hardware design. We evaluate on the smallest Mamba model (130M), as quantization is widely used for edge deployment where smaller models are preferred.  We test with both official Quamba weights and our generated weights using different percentile calibrations. Following Quamba, we report average accuracy of 5 runs on each dataset, with a total of 6 datasets evaluated.

We present our results compared to Quamba in \tab{table:datasetresults}, while the reported results \cite{chiangquamba} serve only as a reference.      On LAMBADA, Quamba-SE achieves up to +2.33\% improvement (Calib. 99.999\%).  The highest gain of single dataset reaches +2.68\% on WinoGrande.

Quamba-SE consistently outperforms Quamba across all calibration settings, with +0.22\%, +0.59\%, +0.83\% improvement in average accuracy. Since small models on edge devices are typically tailored for specific use cases, we further report a subset of benchmarks where the model demonstrates particular strength, showing +0.36\%, +0.70\%, and +1.48\% improvement over Quamba.
The largest overall improvement (+1.48\% on Avg.$^\dagger$) occurs with official pretrained weights. Even when Quamba with 99.99\% percentile slightly outperforms reported results in Quamba \cite{chiangquamba}, Quamba-SE still outperforms Quamba.

\subsection{Discussion}
\textbf{Outlier vs Precision Dilemma:} Covering outliers increases the step size for quantization; dropping outliers causes information loss. Our soft-edge keeps the precision for most values while retaining outliers instead of  hard clipping.

\textbf{Latency:} Our soft-edge quantizer adds latency, but quantization  is a minor computation part in the model, compared with heavy matrix operation. And this minor latency is justified by its accuracy gain. Branch prediction, if applied, can further reduce extra latency overhead.

\textbf{Quamba2 and Significance:} Quamba2 shows minor improvement in Mamba1 models (Quamba1 57.9\%→ Quamba2 58.1\% on 1.4B Mamba1). Our Quamba-SE outperforms Quamba1 by 0.83\% on 130M Mamba1. In LLM area, improvements are typically marginal;
+0.83\% is a meaningful gain given the complexity of multi-dataset evaluation. Evaluation on Mamba2 with Quamba2 are left for future work.

\section{Conclusion} We proposed Quamba-SE, a soft-edge quantizer for Mamba activation quantization that preserves outliers with adaptive scales instead of hard clipping. We evaluate on six datasets with different settings on 130M Mamba model and compare with SOTA work published in 2025. Experiments demonstrate +0.22\% to +0.83\% accuracy improvement over Quamba.

As SSMs' unique activation features demand specific quantization beyond standard INT8 operations. While existing methods focus on improvements within CUDA constraints, we show that customized hardware quantizers can achieve better accuracy. Hardware synthesis is left for the future work. This work represents a promising direction---layer-specific and model-specific quantizer design.

\bibliographystyle{IEEEtran}
\bibliography{ref}
\end{document}